\documentclass[sigconf, nonacm]{acmart}

\usepackage{balance}

\usepackage[ruled,linesnumbered]{algorithm2e} 

\usepackage{algorithmic}
\usepackage{float}
\usepackage{listings}
\begin{document}
%
\title{PatrickStar: Parallel Training of Pre-trained Models via Chunk-based Dynamic Memory Management}

\author{Jiarui Fang, Zilin Zhu}
\affiliation{
\institution{Pattern Recognition Center, Wechat AI, Tencent Inc}            
\city{Beijing}
\country{China}                    
}
\email{{jiaruifang,zilinzhu}@tencent.com}          



\author{Shenggui Li}
\affiliation{
\institution{School of Computing, National University of Singapore}            
\country{Singapore}                    
}
\email{lisg@comp.nus.edu.sg}          


\author{Hui Su, Yang Yu, Jie Zhou}
\affiliation{
\institution{Pattern Recognition Center, Wechat AI, Tencent Inc}            
\city{Beijing}
\country{China}                    
}
\email{{aaronsu, josephyu, withtomzhou}@tencent.com}          

\author{Yang You}
\affiliation{
\institution{School of Computing, National University of Singapore}            
\country{Singapore}                    
}
\email{youy@comp.nus.edu.sg}          

\begin{abstract}
The pre-trained model (PTM) is revolutionizing Artificial Intelligence (AI) technology. 
However, the hardware requirement of PTM training is prohibitively high, making it a game for a small proportion of people. 
Therefore, we proposed PatrickStar system to lower the hardware requirements of PTMs and make them accessible to everyone. 
PatrickStar uses the CPU-GPU heterogeneous memory space to store the model data. 
Different from existing works, we organize the model data in memory chunks and dynamically distribute them in the heterogeneous memory. Guided by the runtime memory statistics collected in a warm-up iteration, chunks are orchestrated efficiently in heterogeneous memory and generate lower CPU-GPU data transmission volume and higher bandwidth utilization. 
Symbiosis with the Zero Redundancy Optimizer, PatrickStar scales to multiple GPUs on multiple nodes.
The system can train tasks on bigger models and larger batch sizes, which cannot be accomplished by existing works. 
Experimental results show that PatrickStar extends model scales 2.27 and 2.5 times of DeepSpeed, and consistently exhibits significantly higher execution speed.
PatricStar also successfully runs the 175B GPT3 training task on a 32 GPU cluster.
Our code is publicly available at https://github.com/Tencent/PatrickStar.

\end{abstract}
\maketitle

%

\section{Introduction}
\label{sec:introduction}


The pre-trained models (PTMs) such as BERT~\cite{devlin2018bert}, GPT~\cite{radford2019language, brown2020language}, and ViT~\cite{dosovitskiy2020image} have become a milestone in the field of Artificial Intelligence (AI).
Although successful pre-training efforts exist on models of trillions of parameters~\cite{shoeybi2019megatron, rajbhandari2020zero, rajbhandari2021zero}, it still has a long way to go before making PTM accessible to everyone.
The application of PTM has two phases, the pre-training phase and the fine-tuning phase.
Although the computation operations are the same, there is a huge gap in the accessible hardware quality between the two stages.
The pre-training phase is trained from scratch on a super large-scale dataset with countless iterations and is extremely time-consuming.
Therefore, it is usually conducted on supercomputers using hundreds of GPU nodes connected with high-speed network fabric.
Such high-quality hardware is only affordable for a small proportion of people in the AI community.
In the fine-tuning phase, the pre-trained models are often fine-tuned on much smaller downstream application-specific datasets.
The hardware quality that most people can access in this phase is much lower than the pre-training phase, usually a single node equipped with multiple GPU cards.
Therefore, making the model size on accessible hardware as large as possible is the critical path to the democratization of PTM.
Unfortunately, the existing work pay little attention to bridging the hardware requirement gap between these two stages.

Heterogeneous training~\cite{pudipeddi2020training,ren2021zero,rajbhandari2021zero} is the most promising solution to lower the hardware requirements of fine-tuning phase.
This approach exploits the heterogeneous memory space including GPU memory and CPU memory and only moves data to the required device if necessary.
However, all of these works reported their results on multiple nodes of the DGX-2 supercomputer.
Each node supplies sufficient memory resources, i.e., 8x32GB GPU memory 1.5TB CPU memory and 3.84TB NVMe SSDs. 
The memory configuration far exceeds the average standard in data centers and cloud computing platforms.
The performance is largely compromised when adopting these systems on commonly accessible hardware.
For example, ZeRO-Offload~\cite{ren2021zero}, integrated into the state-of-the-art (SOTA) heterogeneous training system DeepSpeed, reported that its maximal model scale on a 4xV100 GPU DGX-2H server reaches 30B (Billion) parameters and obtains a throughput of 30 Tflops per GPU. 
However, on a 4xV100 GPU with a 240 GB DRAM CPU server, its maximal model scale can only reach 6B parameters.
In addition, the computing efficiency of Zero-Offload is not optimal.
For the 10B model on 4xV100 DGX-2H server, ZeRO-Offload only utilizes 25\% of the maximal computing power.

We observe two types of tensors managed during PTM training: the model data consists of parameters, gradients, and optimizer states whose footprints are related by the model structure definition;
and the non-model data consists of the intermediate tensors generated by operators.
The non-model data dynamically change according to the configuration of training tasks, such as batch size.
Model data and non-model data compete for GPU memory with each other.
Without considering non-model data volume changing inside an iteration, the existing solutions~\cite{ren2021zero} statically partition the model data between CPU and GPU memory, and their memory layout is constant to various training configurations.
Such a static partition strategy leads to several problems. 
First, the system will crash when the GPU or the CPU memory is insufficient for its corresponding model data requirements, even if there is still memory available on the other devices at the time.
Second, communication is inefficient when data transferring among different memory spaces in the granularity of the tensor, and CPU-GPU communication volume can be reduced when placing model data on the target computing device in advance.

PatrickStar overcomes these shortcomings by managing model data in a chunck manner to use heterogeneous memory more efficiently. 
We organize the model data tensors in \textbf{chunks} which are blocks of continuous memory of the same element size. 
The distribution of chunks in heterogeneous memory space is dynamically orchestrated during training more timely according to their tensor states.
Through reusing chunks that do not coexist, PatrickStar also further lowers the memory footprint of model data than SOTA (state-of-the-art) solutions.
We use a warm-up iteration to collect the statistics of available GPU memory for model data at runtime.
An efficient chunk eviction strategy and device-aware operator placement strategy are designed to reduce the CPU-GPU data movement volume.
Chunk-based memory management can be efficiently symbiotic with data parallelism using Zero Redundancy Optimizer~\cite{rajbhandari2020zero} through the collective inter-GPU communication of chunks.

Our main contributions include:
\begin{itemize}
\item
We build an innovative DNN training system from scratch called PatrickStar, which consists of two modules: chunk-based memory management and dynamic memory management.
\item
The chunk-based memory management is naturally symbiotic with Zero Redundancy Optimizer data parallelism. The chunk-based communication pattern leads to higher CPU-GPU and inter-GPU bandwidth utilization. 
\item
The dynamic memory management can improve memory efficiency and reduce CPU-GPU communication overhead.
\item
We evaluate our system on two GPU clusters, an 8x V100 GPUs 240 GB memory node and an 8x A100 1TB memory node. 
PatrickStar trains GPT-like models 2.27 and 2.5 times the maximal model scale of DeepSpeed and trains faster under the same model scale.
\item
PatrickStar has higher computing and memory efficiency than DeepSpeed and achieves superlinear scalability on 8x GPUs.
\end{itemize}


\section{Backgrounds of PTM}
\label{sec:background}


Nowadays, PTMs use transformer~\cite{vaswani2017attention} structures due to their superior performance.
And to speed up convergence, adaptive gradient-based optimizer ADAM~\cite{kingma2014adam} are the de facto choice in transformer model training.
Taking advantage of the tensor-core component on GPUs, PTM training is usually conducted in a mixed-precision way, which requires that the parameters and gradients of the model use the half-precision floating-point (fp16) format in {forward propagation (FWD) and backward propagation (BWD)}, and the single-precision floating-point (fp32) format during the parameter updating~\cite{micikevicius2017mixed}.
We list the types of tensors used in PTM training.
\textbf{Param fp16}: the model parameters of fp16 type used in FWD and BWD.
\textbf{Grads fp16}: the gradients of param fp16, generated in BWD.
\textbf{Optimizer states (OS)}: state parameters required by ADAM optimizer, including momentum fp32 and variance fp32 and param fp32.
\textbf{Activations}: intermediate output tensors of the operators, also known as feature maps, produced in FWD and BWD.
\textbf{Temporary data}: 
buffers used during operator computation and memory overhead of the software framework.
We refer to activations and temporary data as \textbf{non-model data}, and the others as \textbf{model data}.


\begin{figure}[ht!]
\centering
\includegraphics[width=0.45\textwidth]{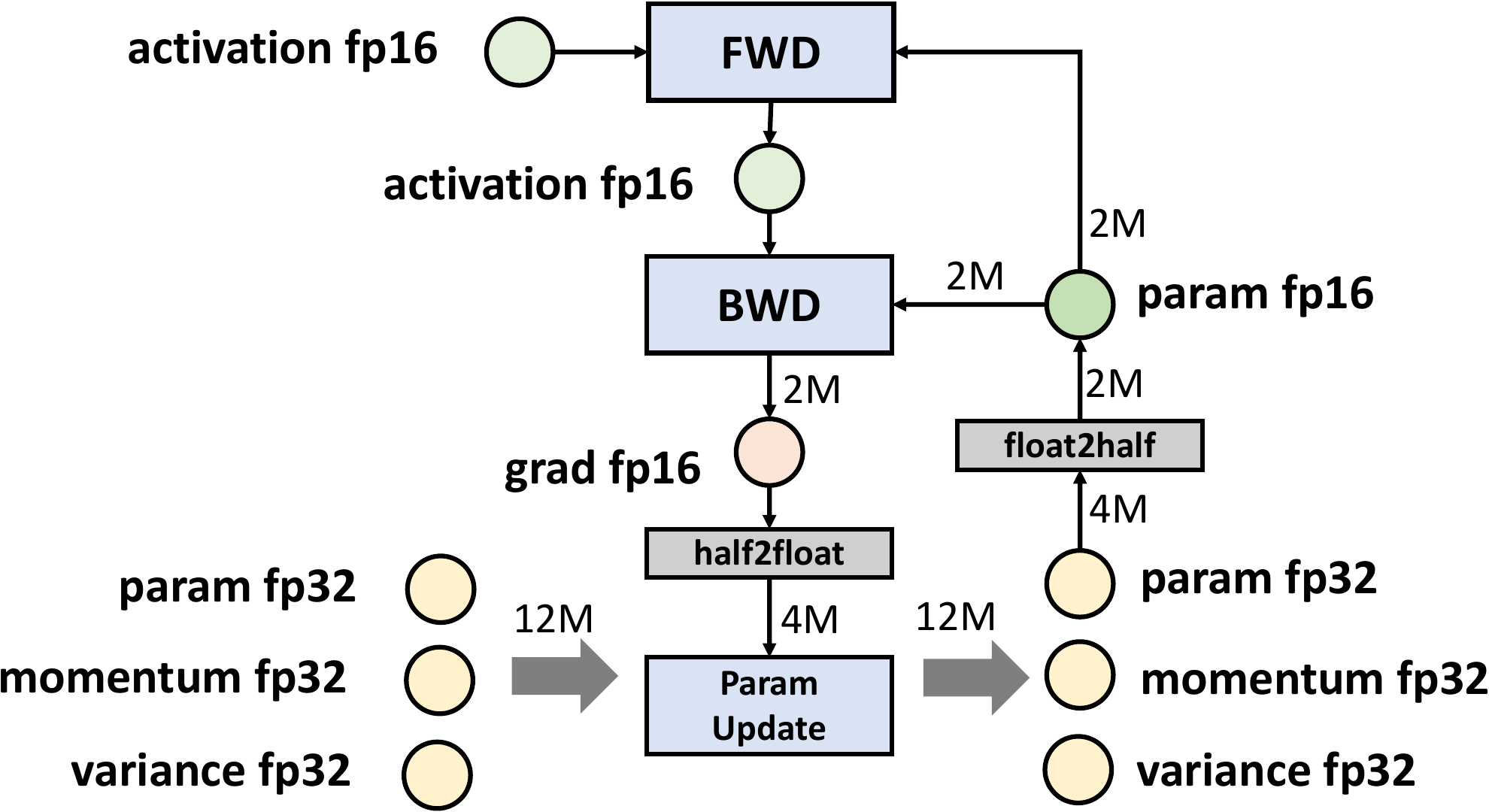}
\caption{Workflow of Neural Network Training (From ~\cite{ren2021zero})}
\label{fig:background-crop}
\end{figure}

The PTM training is viewed as a directed graph. As shown in Fig.~\ref{fig:background-crop}, the circular nodes represent tensor sets (param fp16, grad fp16, param fp32, momentum fp32, variance fp32), and rectangular nodes represent operator sets, including FWD and BWD, Parameter Updating.
Edges of the graph represent the data flow, and the weight of the edge is the overall amount of data (in bytes), assuming the model has $M$ parameters.
Let us first theoretically estimate the memory requirements of PTM training.
For a $M$-parameter model, the FWD and BWD require $4M$ bytes space for param fp16 and grad fp16 tensors.
For the Param Updating, 4$M$ (momentum fp32) + 4$M$ (variance fp32) + 4$M$ (param fp32) = 12$M$ bytes are required.
The updated param fp32 is converted to param fp16 to participate in the computation of the next iteration.
In addition to grad fp16, the overall training process requires 2+2+14=18$M$ bytes of memory space.
For a GPT model with 2 billion (2B) parameters, 36 GB of GPU memory is required, more than 32 GB of V100 overall GPU memory.
And we have not taken the memory usage of non-model data during the training process into account yet.
The GPU memory is often more scarce than expected.

\section{Related Works}
\label{sec:related_work}
Here list the three major lines of work of PTM training.
Some of them are often used together.

\textbf{Parallel Training}: The most popular parallel training method is data parallelism~\cite{dean2012large, pytorchdist} (DP),
which keeps a copy of the model on each GPU and partitions inputs and activations among multiple GPUs.
DP fits the entire model data in the GPU memory, which is no longer valid in the PTM training situation.
Parallel training PTM, therefore, has to partition the model data across multiple GPUs.
A set of works are proposed to adapt DP on PTM training.
The Zero Redundancy Optimizer(ZeRO-DP)~\cite{rajbhandari2020zero} eliminates memory redundancies of model data by partitioning them in layers among multiple GPUs,
therefore using the total aggregate GPU memory capacity of a cluster.
ZeRO-DP applies extra broadcast communications to collect parameters on remote devices.

Model parallelism (MP)~\cite{shoeybi2019megatron,shazeer2018mesh,li2021sequence, xu2021efficient, wang202125dimensional, bian2021maximizing, bian2021colossal} is also revisited in PTM training recently.
MP splits the multi-dimensional tensors of model data along one or more specific dimensions and then distributes them on multiple devices.
In this case, the activations have to be accommodated on each device redundantly.
MP has poor computational efficiency than DP due to additional communication of activations in both FWD and BWD.
As for memory efficiency, MP reduces the model data footprint proportional to the number of workers while not reducing the non-model data footprint.
In addition, the computing and communication pattern of model parallelism are specific to model architecture, which will induce a great initial implementation effort for different models.

Pipeline Parallelism (PP) ~\cite{huang2019gpipe,narayanan2019pipedream,li2021terapipe} is another approach for distributed training that divides the model's layers into stages that the model can process in a pipeline manner.
As one stage completes the FWD or BWD for a micro-batch, the activation tensors are communicated to the next stage, and the gradients of activations are passed to the previous stage in the pipeline.
Although PP communicates less volume than the DP and MP, it has pipeline starting overheads. Another challenge in PP is load balancing, as the different layers have different amounts of parameters. Thus, it requires a careful layer dividing policy to avoid memory bottlenecks on a certain device.  


\textbf{Heterogeneous Training}: 
Different from distributing the model data on the homogeneous GPU memory space,
heterogeneous training utilizes heterogeneous memory space consisting of both CPU and GPU memory.
L2L~\cite{pudipeddi2020training} accommodates the model data in CPU by default, and the GPU memory is populated only with the executing layer’s model data at any given moment in training.
However, it leads to frequent CPU-GPU data movement, in units of tensors.
Moreover, L2L is implemented on one GPU system.
ZeRO-Offload~\cite{ren2021zero} advanced the ZeRO-DP by offloading the OS data and grad fp16 on CPU and executes ADAM on CPU and transfered grad fp16 and param fp16 tensors between CPU and GPU during training.
Its design details will be elaborated in Section~\ref{sec:motivations}.
ZeRO-Infinity~\cite{rajbhandari2021zero} improves implementation of ZeRO-Offload and extends the offloading method to NVMe memory.

\textbf{Activation Checkpointing and Offloading}: 
In addition to reducing GPU model data, a set of work reduces the GPU memory consumption of activations.
The actitvation checkpointing, a.k.a  activation rematerization, first proposed in work~\cite{chen2016training} and widely investigated in work~\cite{gruslys2016memory,herrmann2019optimal,jain2019checkmate} to trade off activation memory with more computation and is implemented in PyTorch~\cite{paszke2019pytorch}.
Activation offloading~\cite{rhu2016vdnn,shriram2019dynamic,beaumont2020optimal, beaumont2021efficient} is an alternative approach that consists in offloading some of forward activations from GPU to CPU.
Work~\cite{peng2020capuchin} also combines offloading and checkpointing techniques based on dynamic tensor access pattern tracked at runtime.

DeepSpeed~\cite{deepspeedcode} is currently the SOTA PTM training software solution.
For the very-large-scale training tasks, it combines ZeRO-DP, MP, PP as 3D parallelism, trades off these parallelisms and maps workload onto hundreds of GPUs by restricting MP and DP among GPUs inside a node and PP among nodes.
Fore finetuning tasks, it has been proved that ZeRO-offload is the most effective technique to improve the model scale with limited hardware budget.


\section{Motivations}
\label{sec:motivations}
PTM goes through two stages: pre-training and fine-tuning.
In the pre-training stage, AI supercomputers are widely applied.
Parallel training work of Sec.~\ref{sec:related_work} is mainly focused on this stage.
The hardware settings of related works~\cite{shoeybi2019megatron,rajbhandari2021zero} are the latest AI supercomputer with 32 high-end DGX-2H nodes.
Megatron-LM's 8.3 B model~\cite{shoeybi2019megatron} and ZeRO-Infinity-3D's 500 B
model~\cite{rajbhandari2021zero} 
~\footnote{The 32 DGX-2H nodes have a total of 512 Tesla V100 GPUs. Each node has a 1.5 TB CPU memory. The intra-node/inter-node communication bandwidth is 300 GBps/100 GBps}.
While expanding the pre-training model scale is only a game for a small proportion of people,
the fine-tuning stage must involve most people playing with their datasets.
AI developers commonly use the equipment from data centers or self-purchased small-scale servers.
Their computing scale, storage capability are lower than AI supercomputers, i.e., the CPU memory in the data center devices is far below the TeraByte level and is much less than that of a DGX-2 node.
To our knowledge, most DL offline tasks in a GPU data center of an industrial AI lab are running on a scale of fewer than 8 GPUs.
Besides improving hardware qualities, developing a software system to bridge the computing power gap between the pre-training and fine-tuning is the key to democratizing PTM.

The most successful effort to facilitate PTM fine-tuning is the heterogeneous training approach.
DeepSpeed~\cite{deepspeedcode} integrated with optimizations of ZeRO-Offload and Zero-DP~\cite{ren2021zero, rajbhandari2020zero} trains a 13B model on a single V100 of a DGX-2H server.
They statically manage model data in heterogeneous memory space such that the param fp16 data are stored in GPU while the grad fp16 and OS data are stored in CPU.
The param fp16 and grad fp16 data, overall $4M$ bytes, are moved between CPU and GPU during each iteration.
It is worth noting that the experiment environment of ZeRO-Offload has 1.5TB CPU memory, which undertakes most of the model data storage.
However, when decreasing the amount of CPU memory, the maximum model scale drops a lot, e.g., deploying it with 240 GB CPU memory, the maximum model scale is lowered to 4B.
Moreover, its computing efficiency decreases as the model scale increases, pointing to the direction for further improvements.
For example, 47 Tflops can be achieved on a 1B model whereas only 33 Tflops achievable on a 4B model.

\begin{figure}[ht!]
\centering
\includegraphics[width=0.5\textwidth]{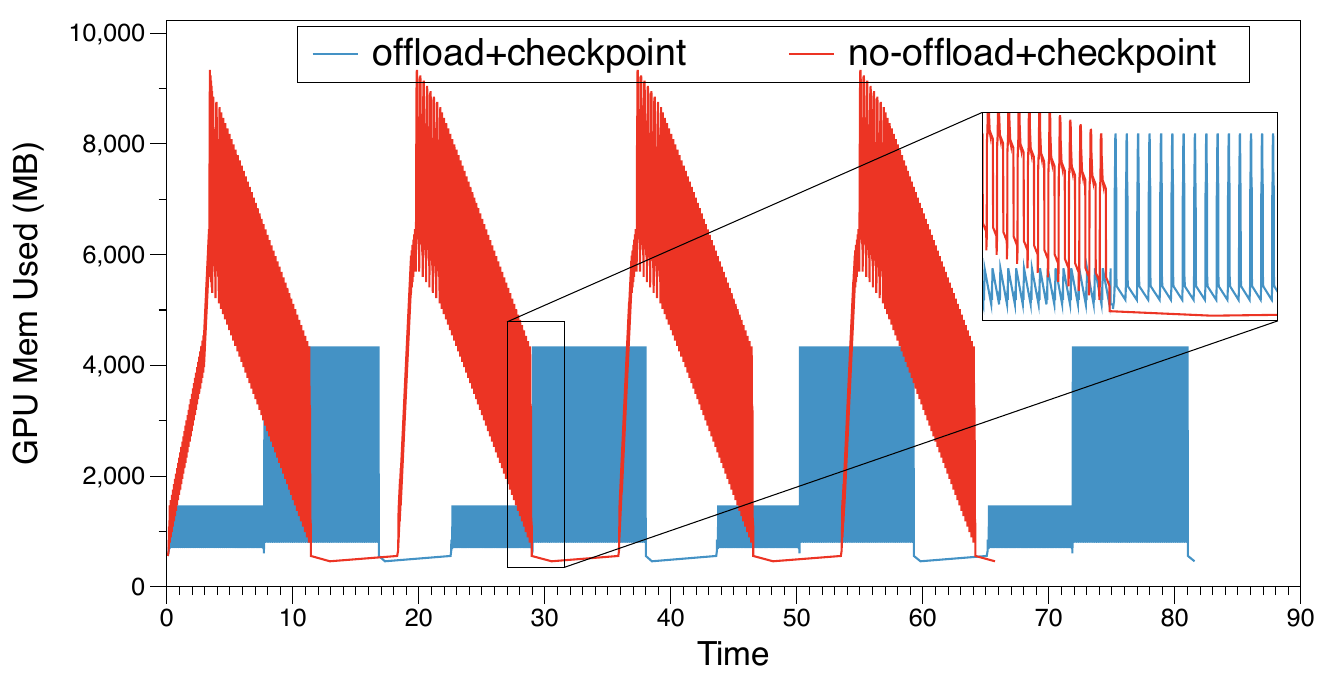}
\caption{GPU memory footprint of non-model data during 4 iterations of a 6B GPT-like model training with PyTorch using activation checkpointing and offloading. The batch size is 16.}
\label{fig:SystemGpuMemUsed}
\end{figure}


ZeRO-Offload is relatively fragile and inefficient on lower quality computing hardware.
During FWD and BWD, GPU memory has to accommodate peak non-model data in addition to overall param fp16.
During param updating (ADAM), the CPU has to accommodate all of the model data.
The memory of model-data depends on the configuration of the model structure, which we call \textit{model-related configuration}.
However, the memory of non-model data is not only related to model-related configuration, but also related to the \textit{task-related configuration}, e.g. batch size, the activation optimization scheme.
It should be noted that the memory consumption of non-model cannot be ignored.
As shown in the {Figure~\ref{fig:SystemGpuMemUsed}}, even if the checkpointing and offloading optimization are used, 
there is still a peak memory consumption of nearly 5GB for this task, and activation offloading has been slowing down the overall training speed.
Using the above static partition strategy, the model-related configuration is strongly affected by task-related configuration.
Once the param fp16 volume exceeds the size of GPU memory limitation subtracted by peak non-model volume, 
the system will crash with OOM, even if the memory shortage only lasts a short period around the non-model memory peaking moment.
Another issue comes from bandwidth utilization. 
ZeRO-Offload and L2L~\cite{pudipeddi2020training} transfers data among different memory spaces in granularity of tensor.
Bandwidth maybe wasted when moving tensors of small message sizes~\cite{li2019evaluating}. 
Bandiwdth underutilization therefore results in lower transmission efficiency and longer execution time.



The above phenomena motivate us toward a new heterogeneous training system design that focuses on 1) improving communication bandwidth utilization efficiency, and 2) optimizing memory utilization.
First, in replacement of small tensor transmission, we adopt a large granularity memory chunk transmission method to optimize both the CPU-GPU and inter-GPU transmitting bandwidth, which has traditionally been the bottle-neck of heterogeneous training systems~\cite{pudipeddi2020training, ren2021zero}. According to the PyTorch Team's observation~\cite{pytorchdist}, \textit{communications are more efficient on large tensors (to saturate the bandwidth, it is suggested to transmit over 20-M elements each time)}, but with ZeRO-DP, 97\% of the parameter tensors are less than 2.6-M when training a 4B GPT~\cite{radford2019language} model on 8 GPUs. Moreover, introducing MP further reduce the model tensor sizes and lower the bandwidth utilization. To address this issue, we came up with chunk-based memory management as inspired by PyTorch-DPP~\cite{pytorchdist}, which uses \textit{Gradient Bucketing} to concat gradient tensors into a memory chunk to improve the inter-GPU bandwidth utilization. 
Second, instead of statically splitting the model data into GPU and CPU, we dynamically arrange them into GPU and CPU before the execution of each operator. Briefly, when there are too much non-model data, the param fp16 tensors not used in the upcoming operators can be moved to the CPU, so that the memory use efficiency is optimized and the trainable model scale is enlarged.

\section{Design Overview}

\begin{figure}[ht!]
\centering
\includegraphics[width=0.45\textwidth]{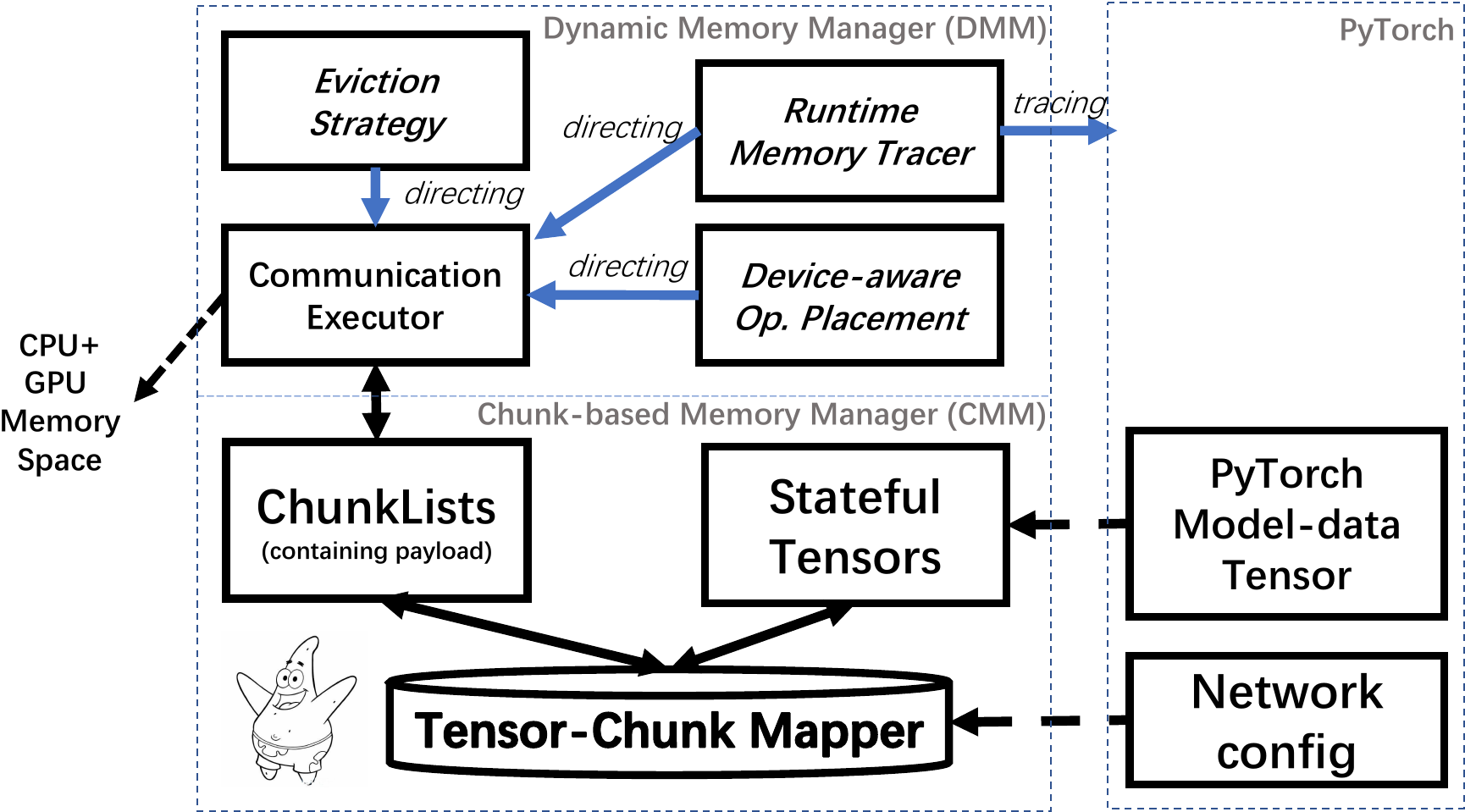}
\caption{The Software Architect of PatrickStar.}
\label{fig:design_overview}
\end{figure}

As shown in the Fig.~\ref{fig:design_overview}, PatrickStar is the middleware of PyTorch and heterogeneous memory space.
The software consists of two modules: a \textbf{Chunk-based Memory Manager (CMM)} and a \textbf{Dynamic Memory Manager (DMM)}.
CMM organizes Pytorch tensors into chunks to improve communication bandwidth (elaborated in Sec.~\ref{sec:cmm}).
DMM dynamically arrange model data layout in CPU and GPU before operation executions, which can reduce CPU-GPU communication volume and thus increase the model-scale (elaborated in Sec.~\ref{sec:opt}).
The DMM module can work independently from PatrickStar and be applied to other PTM training frameworks.


\section{Chunk-based Memory Management}
\label{sec:cmm}
\begin{figure}[ht!]
\centering
\includegraphics[width=0.5\textwidth]{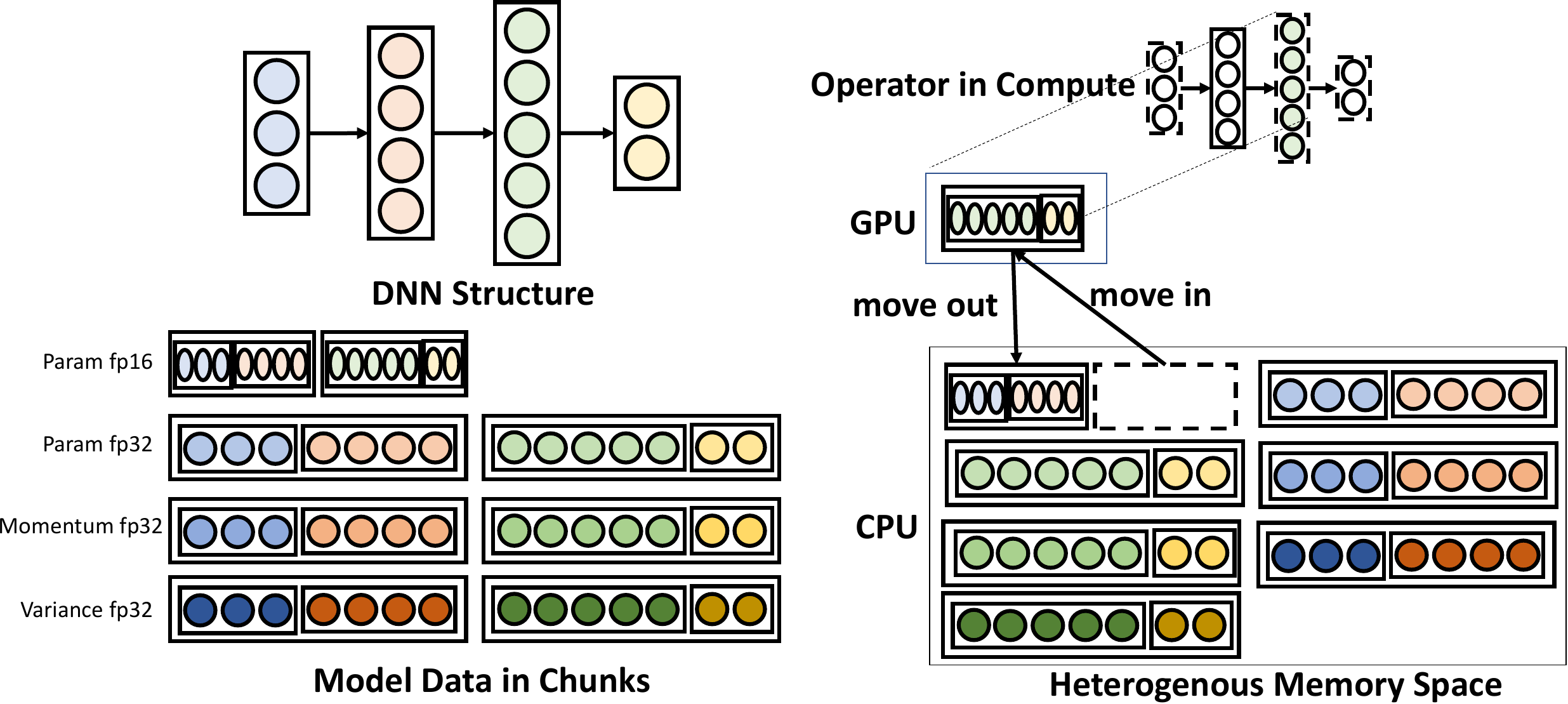}
\caption{The working mechanism of CMM.}
\label{fig:one_gpu_overview}
\end{figure}

The mechanism of Chunk-based Memory Management on a single GPU is shown in Fig.~\ref{fig:one_gpu_overview}.
PatrickStar improves the bandwidth utilization of existing heterogeneous training by managing the model data into chunks (details in Section~\ref{sec:1_gpu}).
The circles in the figure represent the parameter elements, and they are arranged in memory chunks.
When the GPU computation of an operator is triggered (the light-green part of the DNN in right figure), 
PatrickStar moves the required chunks to GPU memory.
Moreover, it is scalable to multiple GPUs combining with ZeRO-DP~\cite{rajbhandari2020zero} (details in Section~\ref{sec:scale_m_gpu}).

\subsection{Design on a Single GPU}
\label{sec:1_gpu}


Before training, the CMM builds the \textit{Tensor-Chunk-Mapper} by constructing the mapping schema between tensor and chunk based on the neural network structure.
During training, it takes over the memory access of the PyTorch tensor and directs it to the memory piece in \textit{ChunkLists}.


\subsubsection{Preprocessing Stage}
\label{sec:one_gpu_preprocess}
We propose an efficient chunk-tensor mapping method with three features: 
1) increase the locality of tensor access.
2) reduce peak memory usage.
3) be parallel-friendly.
Our mapping schema is derived from the following procedure.
Chunks are divided into four types according to types of tensors in model data, i.e., param fp16 list, param fp32 list, momentum list, and variance list, overall 14$M$ bytes ($M$ is parameter number).
Chunks contain the same elements, so different chunks in the same type can reuse the same memory space.
Especially, PatrickStar does not allocate a grad fp16 list.
The grad fp16 tensors can reuse chunk space of the param fp16 list
since we eliminate the dependency of grad fp16 tensors on param fp16 tensors.
PatrickStar shrinks the memory footprint compared to ZeRO-Offload~\cite{ren2021zero} whose minimal model-data memory footprint is 18$M$ bytes as mentioned in Section~\ref{sec:background}.
Additionally, it allocates extra GPU memory holding the gradients to be moved to the CPU, while PatrickStar eliminates this overhead.

We first build the param fp16 chunk list in the order of model initialization.
In this way, when accessing a tensor in a chunk, the neighboring tensors used soon have already been in the same chunk.
Therefore, the access locality is well guaranteed.
For param fp32, momentum, and variance tensors, their corresponding chunk lists are mapped in the same way as param fp16.
As a result, the offsets in the chunk list of param fp16, param fp32, momentum, and variance tensors of the same parameter are consistent.
Scaling to multiple processes, four chunk lists are split at the same position. Therefore the chunks used by ADAM are in the local process and avoid cross-process communication.
The mapping schema generates a certain amount of memory fragmentation. 
We use a lightweight tool to search for the best chunk size of minimal fragmentation before training. 
Experimental results show the percentage of fragments is usually below 10\% for common DNN models.
If there exist tensors breaking the locality with irregular access patterns, which are not common in PTM, PatrickStar supports labeling them not managed in chunks.


\begin{table}[ht!]
\centering
\scriptsize
\caption{States of Tensor in PatrickStar.}
\begin{tabular}{|c||c|c|}
\hline
STATE NAME & EXPLANATION & PLACEMENT \\
\hline
\hline
FREE & No payload space & -\\
\hline
COMPUTE & Participate in computation & Computing Device\\
\hline
HOLD & Hold payload & CPU or GPU \\
\hline
HOLD\_AFTER\_FWD & Hold payload after FWD & CPU or GPU\\
\hline
HOLD\_AFTER\_BWD & Hold payload after BWD & CPU or GPU \\
\hline
\end{tabular}
\label{tab:tensor_state}
\end{table}

\subsubsection{Training Stage}
\label{sec:one_gpu_training}
During training, PatrickStar's tensor is stateful as shown in Table~\ref{tab:tensor_state}.
We refer to the last three states as HOLD-like states.
The COMPUTE state indicates the tensor is about to be computed by an operator on a specific computing device, e.g., CPU or GPU.
The HOLD\_like states indicate that the tensor is not involved in computation right now, but its payload must be maintained in memory, either CPU or GPU.
We distinguish between the HOLD state after FWD and BWD, which is crucial for the following algorithm design.

A chunk's location in heterogeneous space is determined by the states of all its tensors. When all tensors of a chunk are in the FREE state, 
the chunk's memory space can be reused by other chunks or released.
If any of the tensors of a chunk is in COMPUTE state, the chunk must be located on the required computing device.
If none of its tensors is in COMPUTE and at least one of its tensors is in a HOLD-like state, the chunk may be in any place of the heterogeneous memory space.

\begin{algorithm}
\scriptsize
\caption{{Access Tensor}}
\label{alg:access_tensor}
\SetKwProg{func}{Function}{}{end}
\func{FetchRemoteChunks(chunk\_id, comp\_dev)}{
    \state \textbf{Require:} \textbf{Chunks}(a vector of chunks)\;
    \state comm\_grp\_ids = get\_comm\_grp(chunk\_id)\;
    \label{alg:access_tensor_comm_grp_id}
    \If{all chunks in \textbf{Chunks}[comm\_grp\_ids] are local} {
        \state return; (Remote chunks of this comm. group are fetched and not released yet.)\
    }
    \label{alg:fetch_remote_chunk_1}
    \state chunk\_list = list[]\;
    \label{alg:chunk_list_begin}
     \For{id in comm\_grp\_ids}{
        \state prepare payload on comp\_dev\ for \textbf{Chunks}[id] and set all tensors of it to HOLD\;
        \label{alg:dist_access_tensor_prepare_payload}
        \state bind \textbf{Chunks}[id] on comp\_dev; (do not move to other devices)\
        \state chunk\_list.append(\textbf{Chunks}[id])\;
     }
     \label{alg:chunk_list_end}
    \state local\_chunk\_id = comm\_grp\_ids[get\_local\_rank()]\;
    \label{alg:local_chunk_id}
    \state \textbf{ALLGATHER}(chunk\_list, \textbf{Chunks}[local\_chunk\_id])\;
    \label{alg:allgather} 
    \For{id in comm\_grp\_ids}{
        \state unbind \textbf{Chunks}[id] from comp\_dev\;
    }
}
\SetKwProg{func}{Function}{}{end}
\func{Access(param: Torch.Parameter, comp\_dev)}{
    \state \textbf{Require:} \textbf{Chunks}(a vector of chunks)\;
    \state chunk\_id = get\_chunk\_id(param)\;
    \label{alg:access_tensor_chunk_id}
    \If{is\_distributed()}{ \label{alg:access_tensor_dist}
        \state FetchRemoteChunk(chunk\_id, comp\_dev))\;
    }
    \state prepare payload on comp\_dev\ for \textbf{Chunks}[chunk\_id]\; \label{alg:access_tensor_prep}
    \state param.data $\leftarrow$ the corresponding memory piece of \textbf{Chunks}[chunk\_id]\;
    \state old\_state = param.ps\_attr.get\_state()\;
    \If{old\_state is FREE}{
        param.data.zero()\_\;
    }
    \state param.ps\_attr.set\_state(COMPUTE)\; \label{alg:access_tensor_state_update}
}
\end{algorithm}

\begin{algorithm}
\scriptsize
\caption{{Release Tensor}}
\label{alg:release_tensor}
\SetKwProg{func}{Function}{}{end}
\func{ReleaseRemoteChunks(chunk\_id, training\_stage, status, is\_allreduce)}{
    \state \textbf{Require:} \textbf{Chunks}(a vector of chunks)\;
    \state comm\_grp\_ids = get\_comm\_grp(chunk\_id)\;
    \state local\_chunk\_id = comm\_grp\_ids[get\_local\_rank()]\;
    \For{id in comm\_grp\_ids}{
        \If{training\_stage is FWD and not all tensor in \textbf{Chunks}[id] is HOLD\_AFTER\_FWD}{
            \state return\;
        }
        \If{training\_stage is BWD and not all tensor in \textbf{Chunks}[id] is HOLD\_AFTER\_BWD}{
            \state return\;
        }
    }
    \If{is\_allreduce}{
        \state chunk\_list = []\;
        \For{id in comm\_grp\_ids}{
            \state prepare \textbf{Chunks}[id] on comp\_dev\;
            \state bind \textbf{Chunks}[id] on comp\_dev\;
            \state chunk\_list append \textbf{Chunks}[id]\;
        }
        \state \textbf{REDUCE\_SCATTER}(\textbf{Chunks}[local\_chunk\_id], chunk\_list,
        AVG)\;
        \For{id in comm\_grp\_ids}{
            \state unbind \textbf{Chunks}[id] from comp\_dev\;
        }
    }
    \For{id in comm\_grp\_ids}{
        \If{i is not local\_chunk\_id}{
            \state release payload of \textbf{Chunks}[id] and set all tensors of it to FREE\;
        }
    }
}
\SetKwProg{func}{Function}{}{end}
\func{Release(param: Torch.Parameter, training\_stage, target\_state)}{
    \state \textbf{Require:} \textbf{Chunks}(a vector of chunks)\;
    \state chunk\_id = get\_chunk\_id(param)\;
    \state param.ps\_attr.set\_state(target\_state)\;
    \If{is\_distributed()}{ \label{alg:release_tensor_dist}
        \state ReleaseRemoteChunks(chunk\_id, training\_stage, status, training\_stage is BWD))\;
    }
    \state param.data $\leftarrow$ dummy tensor
}
\end{algorithm}

The state of param fp16 tensor is set to HOLD after being initialized (randomly initialized or loaded from the pre-train model).
Before operator FWD computing starts, PatrickStar uses Algorithm~\ref{alg:access_tensor} to access the tensors from chunks.
In line~\ref{alg:access_tensor_prep}, the chunk containing the param fp16 to be computed has to be resident on the computing device.
If not, the chunk manager will move the chunk to the computing device from other devices, which may evict a chunk on the computing device if its memory is full.
The details of chunk eviction strategy of chunk manager will be elaborated in Section~\ref{sec:chunk_evict}.
At last, state of the tensor is converted to COMPUTE (Line~\ref{alg:access_tensor_state_update}), which is stored in tensor's attribute  ps\_attr.
Line~\ref{alg:access_tensor_dist} is used for parallel training and will be elaborated Section~\ref{sec:scale_m_gpu}.
After the computing is finished, we release tensor using Algorithm~\ref{alg:release_tensor} by setting training\_stage as FWD and target\_state as HOLD\_AFTER\_FWD.
The tensor states become HOLD\_AFTER\_FWD, so that the tensor can be evicted to other device if necessary.

After all operators of the model finished FWD, the states of all param fp16 tensors are reset to HOLD to ensure the correct execution of BWD.
This is because the activation checkpointing optimization~\cite{chen2016training} will conduct FWD computation between two checkpoints during BWD.
In this way, if the HOLD states generated by FWD and BWD are not distinguished, we cannot know whether all the tensors in the chunk have finished BWD.

\begin{figure}[ht!]
\centering
\includegraphics[width=0.4\textwidth]{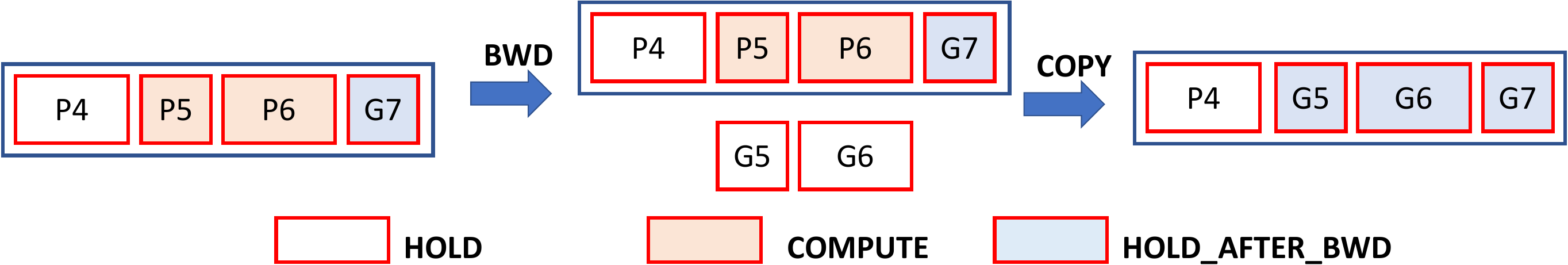}
\caption{Reusing param fp16 chunk with grad fp16. Operator computes on param 5 and 6. P is param and G is grad.}
\label{fig:chunk_reuse}
\end{figure}

For BWD, the inputs of a BWD operator are activations and param fp16 tensors, and the outputs are the gradient of activations and grad fp16 tensors.
Figure~\ref{fig:chunk_reuse} shows how to reuse param fp16 chunk for grad fp16 tensors.
Before the BWD computing, the operator accesses the param fp16 tensors and changes their states to COMPUTE.
During the BWD computing, the generated grad fp16 tensors are allocated on temporary memory space.
After the BWD computing, since param fp16 is no longer needed,  we copy the grad fp16 data from the temporary memory space to the memory space of the corresponding param fp16 tensors and change the tensor state to HOLD\_AFTER\_BWD.
Given that parameters may be shared by multiple operators, we use a reference counter to indicate the last time the tensor is accessed.

\begin{figure}[ht!]
\centering
\includegraphics[width=0.4\textwidth]{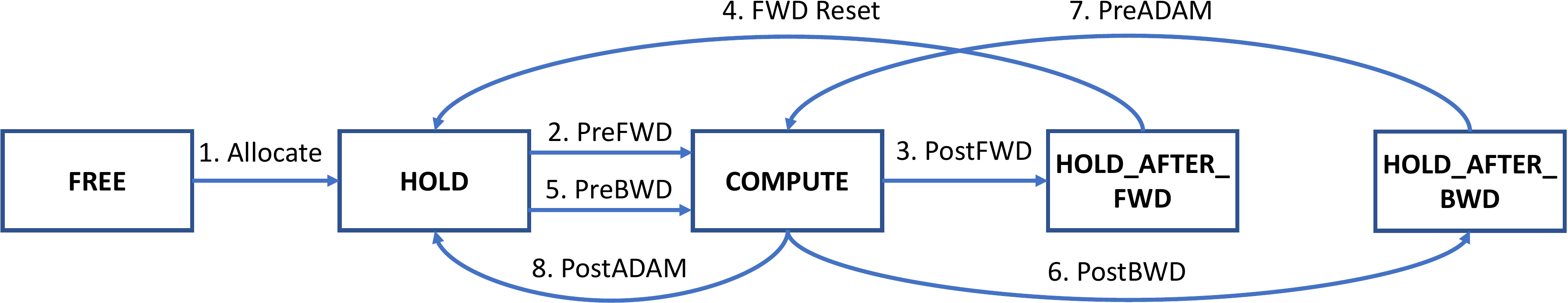}
\caption{The state transition diagram of a param fp16 tensor.}
\label{fig:fsm}
\end{figure}

Before ADAM computing, OS tensors (param fp32, momentum, and variance) are set to COMPUTE.
During computing, grad fp16 chunks are converted to fp32 on the fly to save memory.
After computing, the updated param fp32 tensors and used OS tensors are set to HOLD.
When all tensors in a param fp32 chunk are in HOLD state, param fp32 chunks are copied into the corresponding param fp16 chunk.
Figure~\ref{fig:fsm} shows the state transition of a param fp16 tensor during training.

\subsection{Scaling to Multiple GPUs}
\label{sec:scale_m_gpu}

PatrickStar uses ZeRO-DP~\cite{rajbhandari2020zero} to scale training to multi-GPU via multiple-processing.
Assume the number of processes is $nproc$.
Each process takes care of a single GPU.
In terms of multi-node system, where each node has one CPU and multiple GPUs, the processes inside a node share that one CPU of the node.
Therefore, the local memory space of a process consists of memory of a single GPU and $1/nproc$ of overall CPU memory space.
ZeRO-DP reduces memory requirements by $nproc$ times compared to DP by keeping the $1/nproc$ of the total chunks in its local memory space.
The chunks kept by the process are named \textbf{local chunks}, and the chunks kept by the other process are \textbf{remote chunks}.
A \textbf{communication group} consists of $nproc$ continuous chunks of a chunk list, where each chunk belongs to a different process.

The chunk-based ZeRO-DP works as follows.
Before computing, a process may fetch the remote chunks from the other processes;
After computation of the communication group, the process releases remote chunks.
Distributed logic is added in Algo.~\ref{alg:access_tensor} (line~\ref{alg:access_tensor_dist}) and Algo.~\ref{alg:release_tensor} (line~\ref{alg:release_tensor_dist}).
Owing to the parallel-friendly chunk tensor mapping schema, 
the processes only need to communicate the param fp16 and grad fp16 chunks during the FWD and BWD stages.
The parameter updating requiring the largest amount of data (including momentum, variance, param fp32 and grad fp16), is executed locally.

The \textit{FetchRemoteChunks} of Algorithm~\ref{alg:access_tensor} shows the method of fetching remote chunks.
We identify the communication group of the chunk of the target parameter (line~\ref{alg:access_tensor_comm_grp_id}).
If not yet resident in the local memory of the process (line~\ref{alg:fetch_remote_chunk_1}), 
remote chunks are transmitted to the local via inter-GPU communication.
A list of chunks is prepared in local space, which store the fetched remote chunks after all-gather (line~\ref{alg:chunk_list_begin}-line~\ref{alg:chunk_list_end}).
To avoid the chunk in the list being evicted before all-gather, 
we bind the chunk on the computing device.

\begin{figure}[ht!]
\centering
\includegraphics[width=0.40\textwidth]{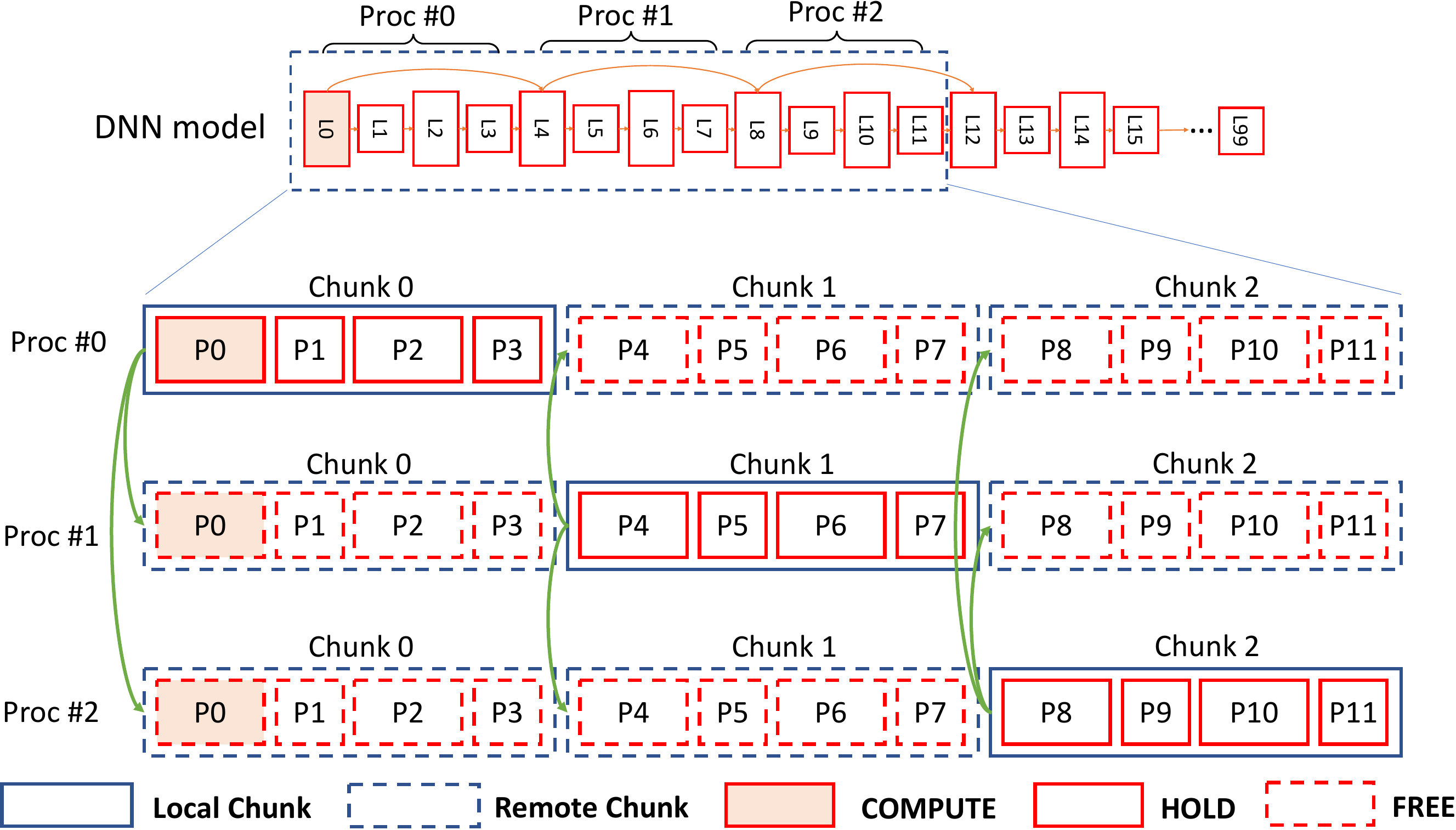}
\caption{Communication pattern for fetching remote param fp16 chunks in before FWD computing of the layer 0.}
\label{fig:fwd_multiple_node}
\end{figure}

Figure ~\ref{fig:fwd_multiple_node} shows a snapshot before the FWD computing on the layer $0$ on 3 GPUs.
The remote chunk is represented by dashed box, and the local chunk is represented by solid box.
The green arrow in the figure indicates the direction of chunk transmission.
In this example, the communication group consists of 3 chunks.
Chunk0 consists of param fp16 tensors of layer 0-3 and belongs to Proc\#0.
Similarly, chunk 1 and chunk 2 belong to Proc\#1 and Proc\#2, respectively. 
Before FWD computing of layer 0, all processes have found remote chunks are not resident in local memory space.
Therefore, all-gather is triggered to fetch remote chunks.
After collecting remote chunks, all processes will have their copy of chunk 0 to chunk 2, and all tensor states in the remote chunks are set to HOLD.
The memory accessing pattern before the BWD computing is the same except in reverse order.

The \textit{ReleaseRemoteChunks} of Algorithm~\ref{alg:release_tensor} shows method of releasing remote chunks.
When the states of all tensors in a communication group are all HOLD\_AFTER\_FWD/BWD after FWD/BWD operator computing finished, 
the tensors in the remote chunks are set to FREE, and the remote chunks are released.
During FWD, we set training\_stage as FWD and is\_allreduce as False.
During BWD, we set training\_stage as BWD and is\_allreduce as True.
A reduce-scatter operation distributes the reduced gradients among processes.

PatrickStar uses collective communication operations for inter-GPU chunk transmitting (allgather and reduce-scatter).
When scaling to multi-node, like over 16 GPUs, 
maintaining a chunk list as the buffer for collective communication consumes excessive memory.
To reduce memory consumption, we sperate a collective operation into $nproc$ serial broadcast and reduction operations so that only one chunk is kept as the communication buffer.
A similar implementation is also used in ZeRO-Offload~\cite{ren2021zero}.

\section{Dynamic Memory Management}
\label{sec:opt}
\begin{figure*}[ht!]
\centering
\includegraphics[width=1\textwidth]{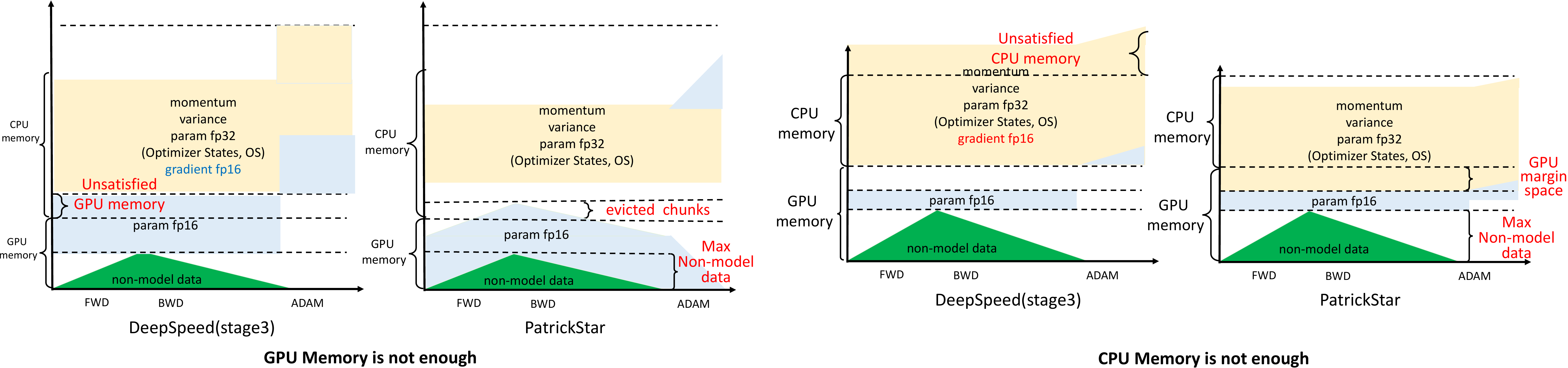}
\caption{PatrickStar can handle situations where DeepSpeed is incompetent.}
\label{fig:adam_hybrid-crop}
\end{figure*}

This section introduces the DMM module as mentioned in Fig.~\ref{fig:design_overview}.
It rearranges model data layout around heterogeneous spaces before operator execution.
Therefore, the system can be used to train a larger-size model.
As shown in the left half of Fig.~\ref{fig:adam_hybrid-crop}.
When the param fp16 and non-model data exceed the GPU memory capacity, data not recently used can be dynamically evicted to the CPU.
Heterogeneous training~\cite{pudipeddi2020training, ren2021zero} usually introduces extra CPU-GPU data movement overhead.
The DMM consists of three innovations that can further reduce the overhead as much as possible.
First, DMM designs a runtime tracer to collect memory statistics (Section~\ref{sec:opt_chunk_prof}).
Second, we can switch the computing device instead of CPU-GPU data moving by a device-aware operator placement optimization (Section~\ref{sec:dao}).
Third, for the data must be moved, PatrickStar uses a smart chunk eviction strategy (Section~\ref{sec:chunk_evict}) to reduce transmission volume.


\subsection{{Runtime Memory Tracer}}
\label{sec:opt_chunk_prof}
DMM obtains the statistics of non-model data memory by a runtime memory tracer.
In the context of PatrickStar, the model-data memory is the same as chunk used memory.
The runtime tracing is necessary since it is hard to get the footprint of non-model data via theoretical estimations.
Even if the shape of each activation tensor is precisely available,
accumulation of all activation tensor sizes is not equal to realistic memory consumption.
First, GPU memory fragmentation and the CUDA context memory is hard to measure before execution.
Second, the dynamic framework PyTorch does not provide an interface to capture the life cycle of every activation tensor before execution. 
Although there is research~\cite{gao2020estimating} on this topic, the method proposed is coupled with the memory allocator details of the target frameworks.
What further complicates the theoretical estimations is that activation tensors whose life-cycle not overlapping with each other may reuse the same memory space~\cite{pisarchyk2020efficient, lee2019device}.
Third, the estimation method is largely affected by activation checkpointing and offloading.

\begin{figure}[H]
\centering
\includegraphics[width=0.42\textwidth]{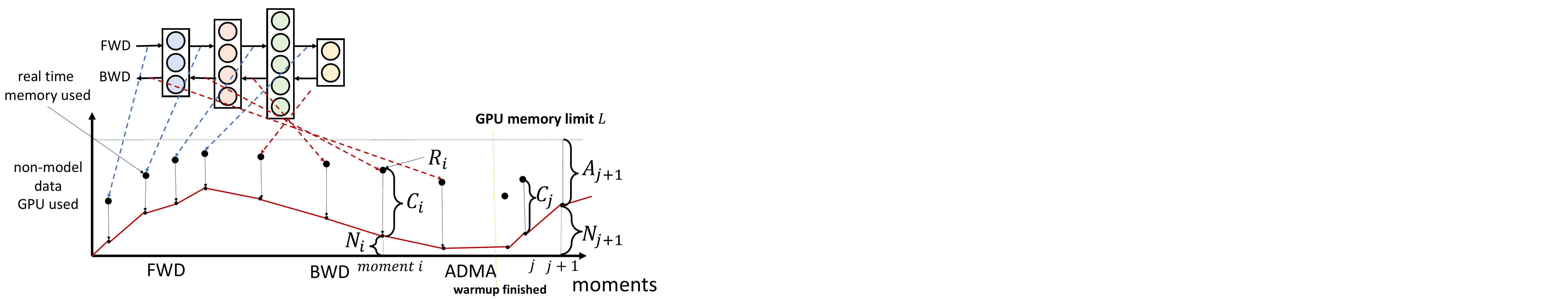}
\caption{Measuring non-data footprint at runtime.}
\label{fig:measuring}
\end{figure}

Fig.~\ref{fig:measuring} shows how we measure non-model data memory footprint in the warm-up iteration.
At the time an operator starts, which is called as \textbf{moment} $i$ here, DMM obtains the real-time GPU memory consumption $R_i$.
DMM can also accurately know the volume of model-data memory $C_i$ in use, therefore, the non-model data memory consumption $N_i$ is figured out with $R_i$-$C_i$.
This traced memory consumption $N_i$ obtained during the warm-up iteration will be used later in other iterations because the computing patters are the same among iterations.
Fig.~\ref{fig:SystemGpuMemUsed} is depicted using the proposed tracer.



\subsection{Device-aware Operator Placement}
\label{sec:dao}
PTM Operators  exhibit various computing and memory access features.
For example, the computing-intensive operator like the \texttt{Linear} must be executed on GPU.
Memory-intensive operators like the element-wise ones in ADAM can be executed on both CPU and GPU.
Note that Memory-intensive operators take up a small part of overall time.
The smart layout of them can improve end-to-end performance by reducing the CPU-GPU communication volume, even if it slightly increases the elapse of their own part.
PatrickStar designs a device-aware placement for optimizer states (OS).
\textbf{GPU margin space} is the remaining space after removing peak non-model data and param fp16 from the overall GPU memory.
The peak non-model data volume is tracked by the memory tracer.
We place as many OS tensors in the margin space as possible.
This way, CPU-GPU communication volume of ADAM is reduced without introducing extra model-data eviction in FWD+BWD.


\subsection{Chunk Eviction Strategy}
\label{sec:chunk_evict}
In a working iteration, the eviction strategy dynamically determines how to move model data between CPU and GPU, depending on runtime statistics of the tracer. As shown in Fig.9, to let the operator execute during the period $j$ to $j$+1, DMM has to prepare at least  max($N_{j+1}$, $N_j$) memory space for non-model data. Consequently, the available memory space for model data $A_{j+1}$ of the next moment $j+1$ is derive by GPU memory limitation $L$-$max$($N_{j+1}$, $N_j$). Assuming the volume of model data not in GPU but required by the current operator is $O_{j}$, then if $C_j$+$O_j$ is larger than $A_{j+1}$, $A_{j+1}$-$C_j$-$O_j$ of data has to be evicted to CPU. In PatrickStar, the chunks prioritized for eviction consist of only HOLD-like tensors.
An inefficient strategy may lead to cache thrashing that the system constantly swap in and out the same chunk.
In the warm-up iteration, the tracer records a list of moments related to each chunk.
A greedy algorithm evicts the longest future reference chunk on this computing device.
It is implemented in $O(log^TC)$ by traversing the moment list of all chunks and binary searching the next moment to be used, where $C$ is the chunk count, and $T$ is the moment count.
Our proposed strategy can be viewed as Belady’s OPT algorithm~\cite{belady1966study, mattson1970evaluation}, which replaces the buffer page with the longest future reference distance.

As for the the warm-up iteration, there is no next moment non-model statistics.
To avoid out of GPU memory, we force only a small proportion, by default 20\%, of GPU memory can be used to store model data.
At this time, it simply evicts chunks in the order of the chunk list.

\subsection{Discussion}
\label{sec:compare_zero}
As shown in Figure~\ref{fig:adam_hybrid-crop}, PatrickStar is competent in two situations where ZeRO-Offload fails.
In the case of GPU not enough (left figure), 
for ZeRO-Offload, when the overall size of param fp16 tensors and peak size of non-model data exceeds GPU memory, unsatisfied GPU memory requirement crashes the system.
With the help of the chunk eviction strategy, PatrickStar makes it work by evicting chunks supposed to be in unsatisfied GPU memory to CPU efficiently.
If CPU memory is not enough (right figure),
there exist unsatisfied CPU memory requirements in ZeRO-Offload.
In PatrickStar, the margin space on GPU can accommodate unsatisfied OS chunks to make the system work smoothly and efficiently.
However, PatrickStar has not yet explore NVMe memory space.
A multi-level cache strategy to extend our proposed eviction strategy will be investigated future.

We emphasize that the DMM module implemented with the above three innovations is decoupled from the CMM module in PatrickStar.
To dynamically move model-data between CPU-GPU in granularity of tensor instead of chunk, the Chunk Eviction Strategy can be adapted to evict the HOLD-like tensor.
We also implemented a system in combination with the ZeRO-DP called \textbf{SpongeBob} using the DMM only but not using chunks to manage tensors.





\section{Evaluation}

\subsection{Evaluation Methodology}
\label{sec:evalmethod}

\textbf{Testbed: }
We conduct our experiments on three GPU clusters. 
The first one is the cloud computing system called \textbf{YARD}.
it is a node configured with 8x 32GB V100 GPU and a 12-core CPU with 240 GB CPU memory.
The second one is called \textbf{SuperPod}~\cite{superpodspecs}.
We use a node consisting of 8x 40GB A100 GPU and 192-core CPU with 1TB CPU memory.
For both clusters, GPUs are connected through NVLink.
The third one is called \textbf{HAI-PCIe}.
It is a node consists of 8x 40GB A100 and 128-core CPU with 512 GB CPU memory.
GPUs are connected with PCI-e.

\textbf{Workloads:} Same as related work, the PTM tasks used for evaluation are GPT-2 like transformer-based models~\cite{hfgpt2}.
The model configuration is shown in in Table~\ref{tab:model_config}.

\begin{table}[ht!]
\scriptsize
\begin{centering}
\caption{{Model Config. (sequence length is 1024, head number is 16)}}
\label{tab:model_config}
\begin{tabular}{ccc||ccclc|c|c|c|cl}
\hline
\#params & \#layer  & hidden dim & \#params & \#layer  & hidden dim\\
\hline
\hline
1,2 B & 20,40 & 2048 & 6,8B & 53, 72 & 3072 \\
\hline
4B & 64 & 2304 & 10, 12 & 78, 90 & 4096\\
\hline
15, 18B & 50, 60 & 4096 & 20, 30B & 25, 37 & 8192 \\
\hline
40, 50, 60B & 50, 62, 75 & 8192 & 68B & 66 & 9126  \\
\hline
\hline
\end{tabular}
\end{centering}
\end{table}

\textbf{Baselines.}
We compare the PatrickStar with PyTorch DDP{~\cite{pytorchdist}} and the state-of-the-art (SOTA) heterogeneous training solution using DeepSpeed~\cite{deepspeedcode} v0.5.3, it has integrated optimizations from the works~\cite{ren2021zero,rajbhandari2020zero,shoeybi2019megatron}.
The configuration of Deepspeed is adopted from the official example.
We use the zero3 stage, indicating using ZeRO-Offload/Infinity optimization.
For all the three software, we apply activation checkpointing.
We pick the best performance with and without activation CPU offloading for PatrickStar and DeepSpeed. Here performance is measured with throughput, which is calculated by dividing the floating-point operation numbers with the average end to end execution time of a iteration.
We update the parameters at every iteration, which is more practical in real scenarios.
We test with various batch sizes per GPU, including 4, 8, 16, 32, 64.

\textbf{Chunk Size Searching.} To minimize the memory fragmentation generated during the mapping processes, we come up with a light-weight method to search for the best chunk size before training is started. 
The chuck size searching script is executed offline on the CPU so that it runs quickly and does not actually allocate memory for parameters. 
This searching method builds the tensor chunk mapping schema by looking for the optimal chuck size that can host the overall model data in CPU+GPU from a size range of 128-M to 512-M with a step of 32-M. 
The optimized chunk sizes of the various models are listed in Table~\ref{tab:bestcs}, where memory fragmentation is minimized (less than 10\%) while chunk memory utilization ratio (UTIL.) is maximized.

\begin{table}[ht!]
\begin{centering}
\caption{Chunk Size Searching Results.}
\label{tab:bestcs}
\begin{tabular}{|c||c|c|c|c|c|c|c|c|c|c|c|c|}
\hline
 & \multicolumn{3}{c|}{YARD} &  \multicolumn{4}{c|}{SuperPod} \\
\hline
Model & 10B	& 15B & 18B & 20B & 40B  & 60B & 68B\\
\hline
\hline
SIZE (M) & 288 & 480 & 312 & 288 & 288 & 416 & 416\\
\hline
UTIL.(\%) & 94.47 & 92.62 & 91.5 & 90.5 & 90.6 & 92.2 & 97.4\\
\hline
\end{tabular}
\end{centering}
\end{table}



\subsection{Experimental Results}
\subsubsection{\textbf{Model Scale}}
\begin{figure}[H]
\centering
\includegraphics[width=0.5\textwidth]{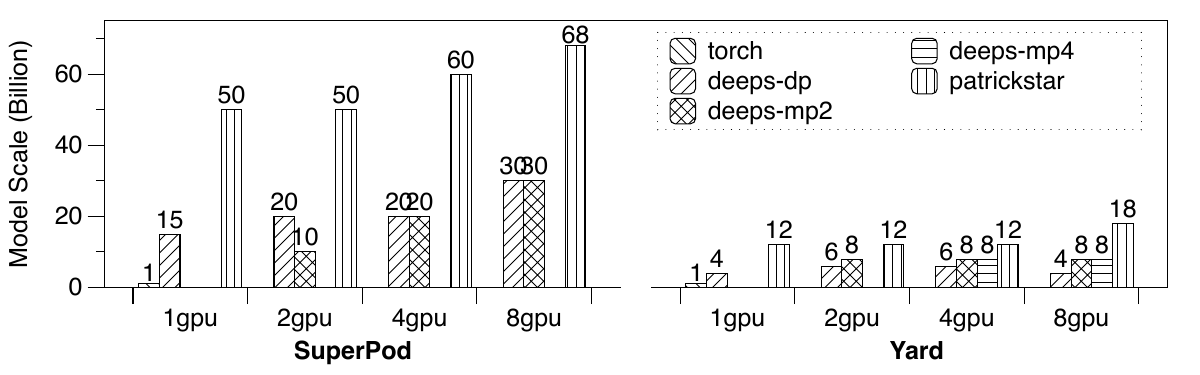}
\caption{{The max model scale on YARD and SuperPod of PyTorch, DeepSpeed and PatrickStar.)}}
\label{fig:model-scale}
\end{figure}

Figure ~\ref{fig:model-scale} reported the model scale of three systems on YARD and SuperPod.
Some entries miss values because of Out-Of-Memory (OOM) or the performance is unable to meet the efficiency bar, which means the throughput is less than 30, 50 Tflops per GPU on YARD and SuperPod, respectively.
The 'Xg' indicates the number of GPUs.
Our cluster requires each GPU to achieve minimal GPU efficiency to meet the energy requirements.

DeepS(eed)-DP is scaling ZeRO-Offload/Infinity~\cite{ren2021zero,rajbhandari2021zero} with ZeRO-DP method.
DeepS(peed)-MP is combined DP with different ways of MP using Megatron-LM~\cite{shoeybi2019megatron}.
For example, deeps-mp2 on 8 GPU indicates using MP is used in 2 GPU group and DP is used in a 4 GPU group.
Since MP requires customized code modification for model definition,
we do not apply MP on PatrickStar.
But they can be used together in the future with extra programming effort.

On a YARD GPU node, PatrickStar improves model scale of DeepSpeed-DP \textbf{3x} from 4B to 12B and improves PyTorch DDP \textbf{12x} from 1B to 12B.
When scaling to 8 GPUs, DeepSpeed-DP can improve model size to 6B on 2 and 4 GPUs. 
However, it decreases the size back to 4B on 8 GPUs.
Compared with DeepSpeed+MP, PatrickStar still improves scale from 8B to 18B (\textbf{2.25x}).
On a SuperPod Node, PatrickStar improves model scale \textbf{3.3x} from 15B to 50B and improves PyTorch \textbf{50x} from 1B to 50B.
When scaling to 8 GPUs, PatrickStar can improve model size from 30B to 68B, \textbf{2.27x} larger than DeepSpeed.


We analyze the impacts of system design on maximal model scale.
PyTorch can only train the cases where model data and non-model data are fitted in the GPU.
When the model size is 2B, the model data reaches 2$\times$18=36 GB.
In addition to non-model data, the GPU memory requirements exceeds the 32 GB V100 and 40 GB A100 capacity.
DeepSpeed can train the case where param fp16 in addition to non-model data can be hosted GPU.
As analyzed in Section~\ref{sec:compare_zero}, even if the overall size of parameters fp16 and non-model data exceed the GPU memory size,
PatrickStar still trains the model smoothly since it can smartly evict current not in use chunks to CPU memory.
{
We observed that the memory efficiency of PatricStar is very high.
On 8 YARD GPUs, a 18B model has 18$\times$14 = 252 GB model data managed in chunks, while the overall CPU+GPU memory for chunk using is 32$\times$20\%$\times$8+240=291.2 GB (20\% comes from model-data GPU memory for warmup from Sec.~\ref{sec:chunk_evict}). 
The memory utilization is 252/291.2=86\%.
On 8 SuperPod GPUs, the memory utilization is 87.5\%.
We believe PatrickStar has reached the utilization limit of CPU+GPU heterogeneous memory space, considering the memory overhead of PyTorch and memory fragmentation of chunks.
}


\subsubsection{\textbf{Computing Throughput On One GPU}}

\begin{figure}[ht!]
\centering
\includegraphics[width=0.5\textwidth]{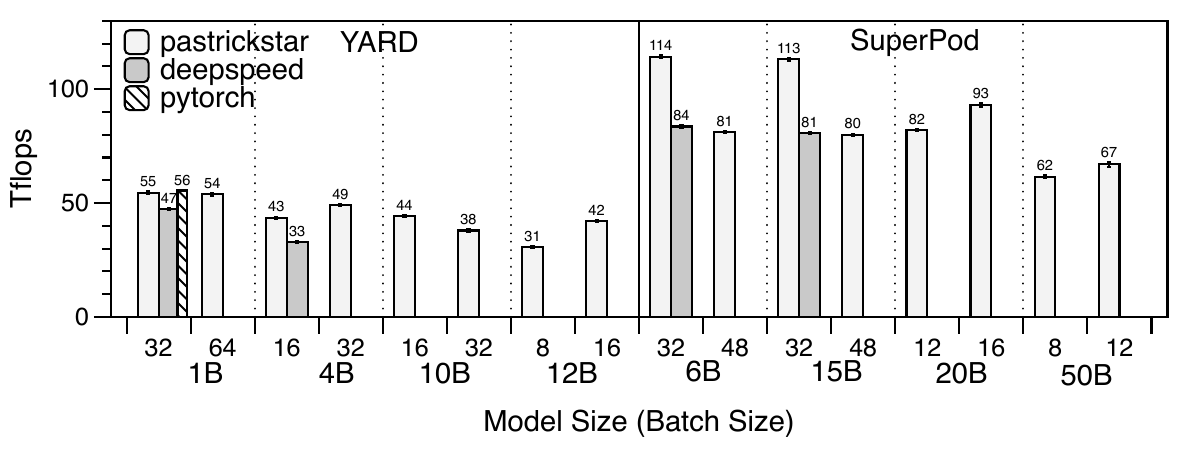}
\caption{
{Training throughput of PyTorch, DeepSpeed and PatrickStar on one GPU in YARD and SuperPod.}}
\label{fig:1gpu_results}
\end{figure}

{Figure~\ref{fig:1gpu_results} shows the performance of three systems on one V100 GPU in YARD and one A100 in SuperPod.
The missing columns indicate DeepSpeed and PyTorch fail in these cases.
The performance of PatrickStar and PyTorch is similar on the 1B model and higher than DeepSpeed.
PatrickStar achieves higher performance and larger batch size than DeepSpeed in all the cases.
Generally, a larger batch size leads to higher GPU utilization.
Note the performance of PatrickStar is lower on batch size 48 than 32 on 6B and 15B model training on SuperPod.
It is because that the large batch size cases use activation offloading and introduce more CPU-GPU data communications.
}

\subsubsection{\textbf{Computing Throughput On Multiple GPUs.}}
\label{sec:res_m_gpu}

\begin{figure}[ht!]
\centering
\includegraphics[width=0.48\textwidth]{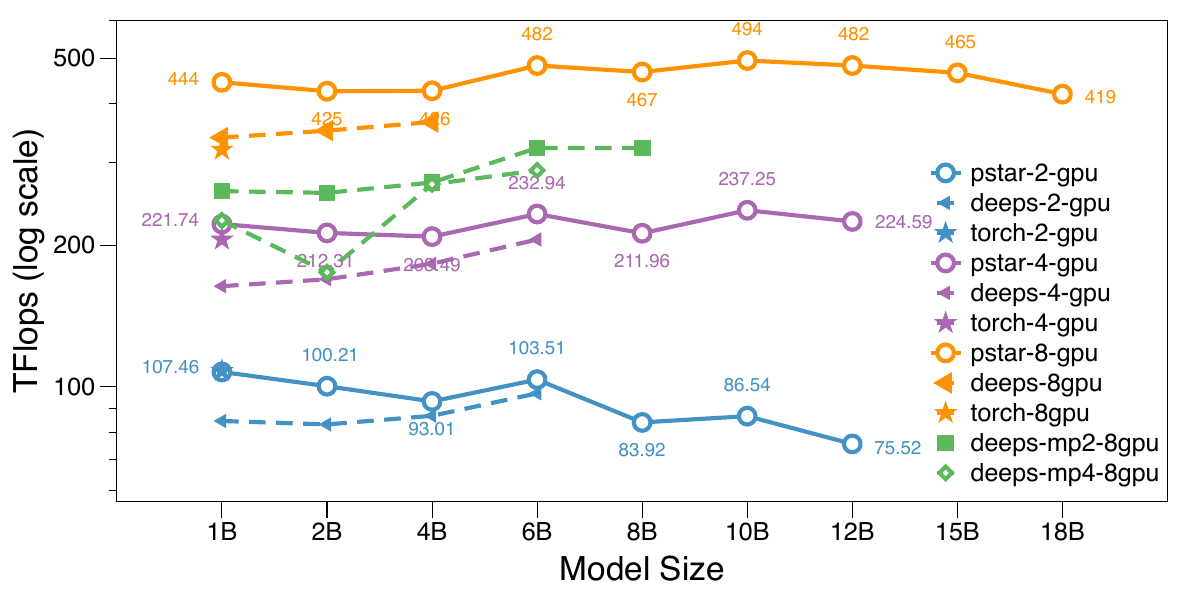}
\caption{
{Training throughput of PyTorch, DeepSpeed and PatrickStar using DP and MP on multiple GPUs in YARD.}}
\label{fig:mgpu_perf}
\end{figure}

\begin{figure}[ht!]
\centering
\includegraphics[width=0.48\textwidth]{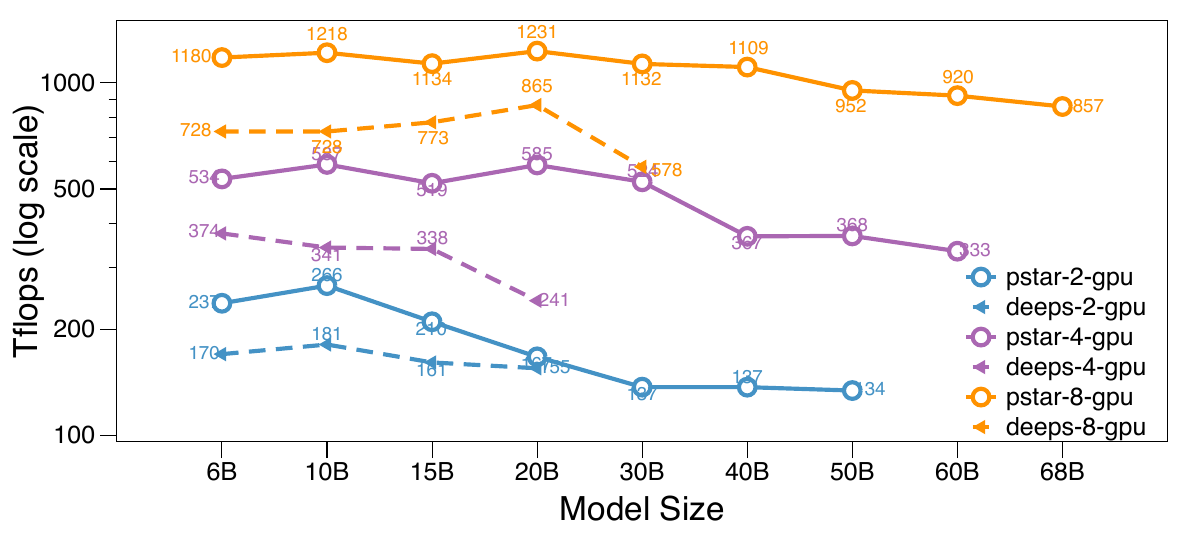}
\caption{
{Training throughput of DeepSpeed and PatrickStar on multiple GPUs in SuperPod.}
}
\label{fig:mgpu_perf_superpod}
\end{figure}

{Figure ~\ref{fig:mgpu_perf} and Figure~\ref{fig:mgpu_perf_superpod} presents the performance of 3 systems on 1, 2, 4, 8 GPUs.}
The y-axis is re-scaled by logarithmic.
The points represent the best results achieved with the best batch size.
{The value around the confidence interval bar indicates average throughput in Tflops}.
The deeps is DeepSpeed-DP, and deeps-mpX is DeepSpeed with X-way MP.

{
Figure ~\ref{fig:mgpu_perf} presents results on YARD.
PatrickStar is 1.37x faster than PyTorch on 8 GPUs and is similar to PyTorch in the 1,2,4 GPU cases for 1B model.
In 8 GPU cases, the maximum batch size of PyTorch is 4 while PatrickStar is 64, 
and small-batch size decreases the efficiency of hardware utilization.
}
On YARD, using the same Zero-DP strategy, PatrickStar is superior to DeepSpeed-DP {in all of the cases} and is the only solution to train model size between 8B and {18B} with DP only.
The improvement is significant {(1.08x-1.47x, on average 1.23x)}, especially for small models.
The reason is also that PatrickStar shrinks CPU-GPU data transmission volume.
PatrickStar does not significantly decrease computational efficiency when increasing the model size.
On 8 GPUs, {the 419 Tflops performance on the 18B model is 94\% of 444 Tflops on 1B.}
This shows that PatrickStar is very robust on the larger model scale with the help of efficient chunk orchestrating.



{
Figure~\ref{fig:mgpu_perf_superpod} presents the results on SuperPod.
MP results are missing because they are always inferior to DP in the same test case.
On SuperPod, the speedup to DeepSpeed is more significant than on YARD (1.07x-2.43x, on average 1.53x).
There is also no significant degradation of performance as the model size scales.
On 8 GPUs, the 857 Tflops performance on the 68B model is 73\% of 1180 Tflops on 6B.
}

\subsubsection{\textbf{Optimization Analysis}}
\label{sec:decom}

\begin{figure}[ht!]
\centering
\includegraphics[width=0.5\textwidth]{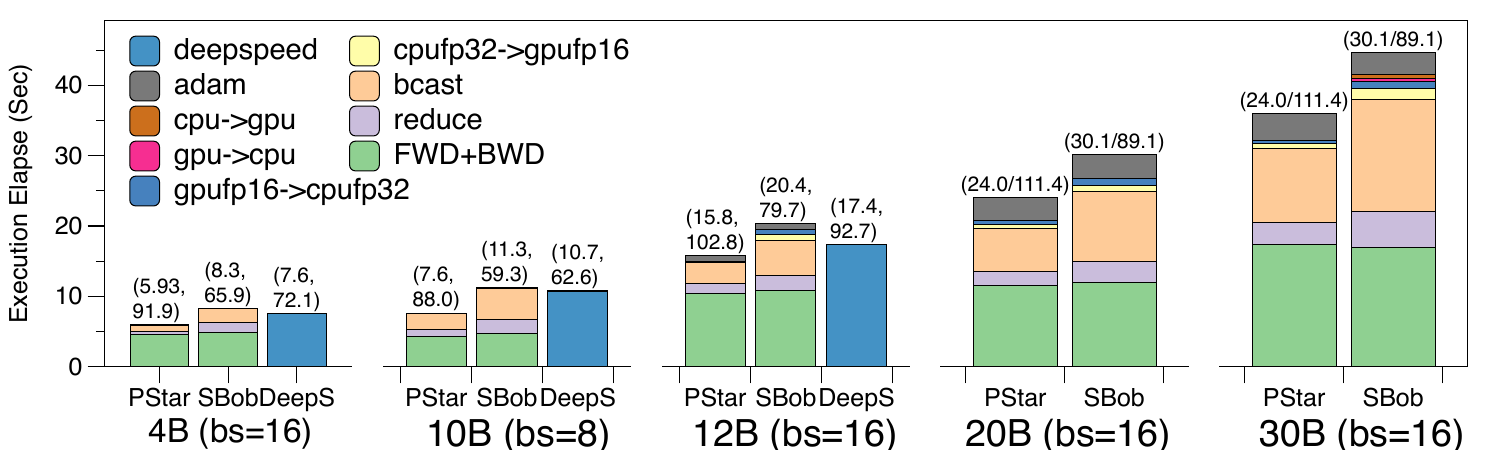}
\caption{
Execution time breakdown of an iteration in PatrickStar, SpongeBob on HAI-PCIe using 8 GPUs. The bs means batch size per GPU. The label shows (Elapse/Tflops per GPU).
}
\label{fig:HPCAI12BTimeDecomp}
\end{figure}

\begin{figure}[ht!]
\centering
\includegraphics[width=0.5\textwidth]{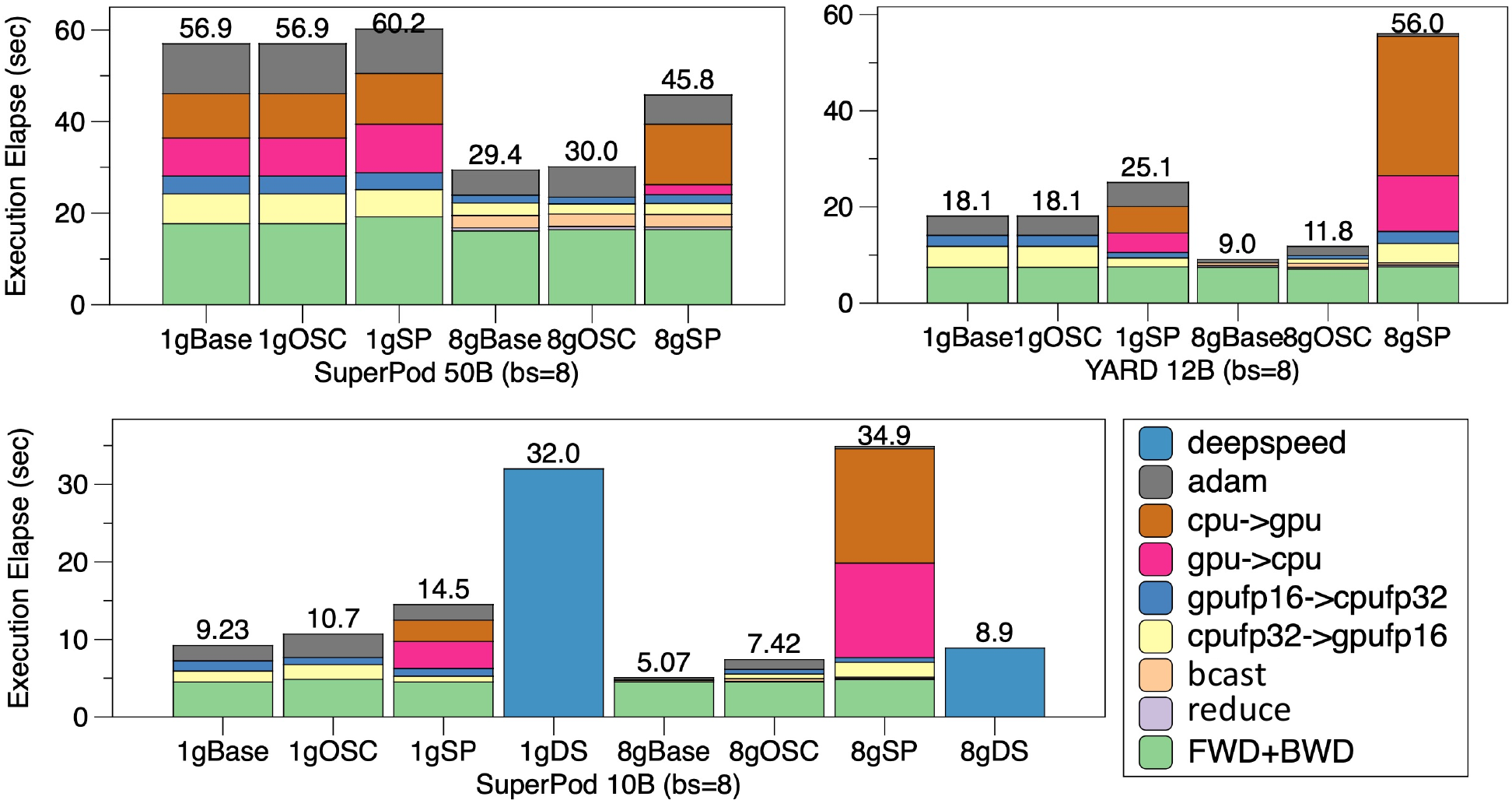}
\caption{
{Execution Time breakdown of a training iteration for different optimization plans on six cases. The labels above the columns show the end-to-end time elapses.}}
\label{fig:timedecomp}
\end{figure}

We highlight the effect of CMM in Fig.~\ref{fig:HPCAI12BTimeDecomp}, where SpongeBob (Sbob) is the system using only DMM as mentioned in Sec.~\ref{sec:compare_zero}.
The figure contains computing parts (ADAM and FWD+BWD), inter-GPU communication parts (all-gather and reduce-scatter) and the chunk moving parts.
The gpu(cpu)$\rightarrow$cpu(gpu) indicates the elapse of CPU-GPU chunk moving for FWD+BWD computing.
The gpufp16(cpufp32)$\rightarrow$cpufp32(gpufp16) indicate the elapse of CPU-GPU chunk moving and accompanying floating-point conversion for ADAM computing. 
DeepSpeed execution time is presented as 'DeepS' in Fig.~\ref{fig:HPCAI12BTimeDecomp} and 'DS' in Fig.~\ref{fig:timedecomp}.


Three systems use the consistent task-related configuration running on HAI-PCIe cluster, which connects GPUs via PCI-e.
DeepSpeed is Out-Of-Memory on the 20B and 30B cases, providing evidence that DMM can effectively improve model scale.
With the help of CMM, PatrickStar exhibits the shortest execution time in all cases.
For the 4B, 10B, 12B cases, it is 1.4x/1.3x, 1.5x/1.4x, 1.3x/1.1x faster than SpongeBob/DeepSpeed.
For the 20B, 30B cases, it is 1.25x and 1.25x faster than SpongeBob,
and has less overhead on inter-GPU and CPU-GPU communication via PCI-e.
The overall communication time of PatrickStar is 54\%, 50\%, 50\%, 51\%, 55\% shorter than SpongeBob in 4B, 10B, 12B, 20B and 30B cases.

Figure~\ref{fig:timedecomp} demonstrates the effect of our proposed optimizations in DMM. Six representative cases were tested using 1 and 8 GPU to train 10B and 50B models on SuperPod and 12B on YARD. In Figure 15, 'XgBase' (X=1 or 8) represents the result of the base PatrickStar on X GPU. 'OSC' represents the cases where OS chunks are fixed on the CPU without Device-aware Operator Placement. 
'SP' represents the cases with static partition where runtime memory tracer is not used and 20\% GPU memory is fixed for model data instead.
Note that this static partition strategy is more robust than DeepSpeed which only works for SuperPod 10B case.

Table~\ref{tab:marginchunks} shows the test cases cover the situation where the all param fp16 can be hosted on the GPU memory and param fp16 has spill to CPU memory.
The last row of the table shows the space size counting in chunk number.
Positive numbers indicate the the number of OS chunks can be held in the margin GPU space.
Negative numbers indicate the number of param fp16 chunks have to be spilled to CPU.

\begin{table}[ht!]
\centering
\caption{The margin space(+) for OS chunks and spilling space(-) for param fp16 chunks.}
\label{tab:marginchunks}
\begin{tabular}{|c||c|c|c|c|c|c|c|c|c|c|c|c|}
\hline
Plan & \multicolumn{2}{|c|}{SPod 10B} &  \multicolumn{2}{c|}{SPod 50B} & \multicolumn{2}{c|}{YARD 12B} \\
\hline
GPU & 1 & 8 & 1 & 8  & 1 & 8  \\
\hline
\hline
margin(+)/spilling(-) & 2 & 6 & -20 & 1 & -1 & 5 \\
\hline
\end{tabular}
\end{table}


{
Runtime memory tracer of Sec.~\ref{sec:opt_chunk_prof} can significantly reduce the volume of CPU-GPU chunk moving during the FWD+BWD.
As shown in the figure, compared with the SP plan, 
the base version almost eliminates the cpu(gpu)$\rightarrow$gpu(cpu) cost.
The resulting end-to-end speedup is impressive.
For example, the 8gBase version is 6.9x faster than 8gSP on 10B SuperPod case.
}

{
Device-aware operator placement of Sec.~\ref{sec:dao} reduces chunk movement and computing time of ADAM.
Compared with OSC, the base version puts part of the OS on the GPU, thereby reducing the overhead of cpufp32(gpufp16)$\rightarrow$gpufp16(cpufp32).
Additionally, the ADAM computation on GPU is faster than on CPU.
It will benefit the cases with enough margin GPU space in Table~\ref{tab:marginchunks}.
For example, the 8gBase version is 1.3x faster than 8gOSC on the 12B YARD case.
}


The chunk-based communication pattern is efficient on SuperPod and YARD.
For SuperPod 10B, 50B and YARD 12B 8 GPU cases, 
The base version's communication overhead (allgather+reduce-scatter) are 5\%, 11\% and 9\%, respectively.
As shown in Table~\ref{tab:nvlinkbwd}, the achieved bandwidth of collective communication in the base version on both clusters is all above 75\% of the saturated bandwidth.

\begin{table}[ht!]
\begin{centering}
\caption{Average Achieved and Saturated Bandwidth on 2 clusters.}
\label{tab:nvlinkbwd}
\begin{tabular}{|c||c|c|c||c|c|c|c|c|c|c|c|c|}
\hline
Bandwidth  & \multicolumn{3}{|c||}{SPod 8 GPU} &  \multicolumn{2}{|c|}{YARD 8 GPU} \\
\hline
(GB/s) & 10B &  50B & Saturated & 12B & Saturated\\
\hline
\hline
Allgather & 160.2 & 158.3 & 208.4 & 90.2 & 112.72 \\
\hline
ReduceScatter & 178.9 & 173.2 & 198.04 & 92.3 & 111.8 \\
\hline
\end{tabular}
\end{centering}
\end{table}

\begin{figure}[ht!]
\centering
\includegraphics[width=0.5\textwidth]{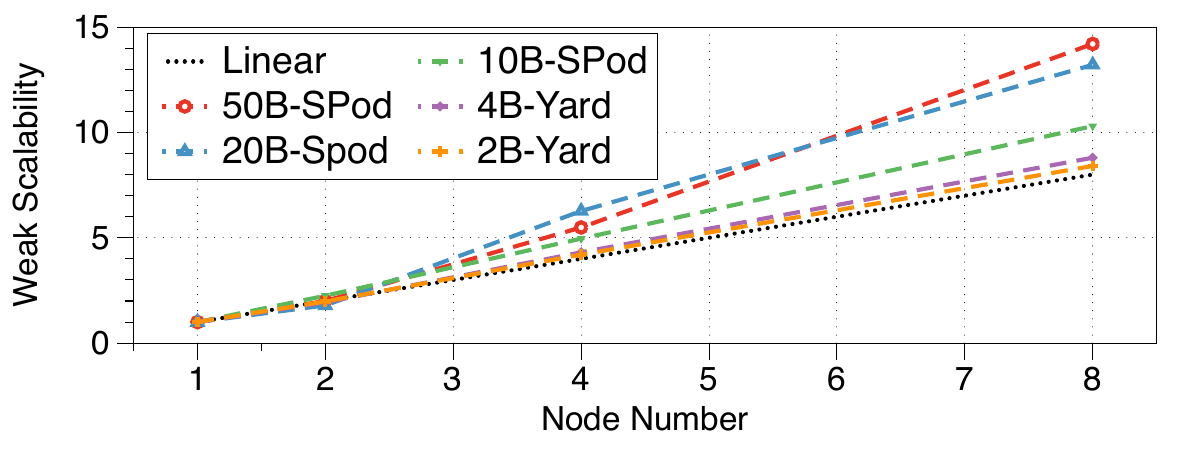}
\caption{
{Scalability of PatrickStar on YARD and SuperPod.}}
\label{fig:scalability}
\end{figure}

PatrickStar also shows superlinear scalability when increasing GPU numbers, as shown in Figure~\ref{fig:scalability}.
Larger models have better scalability, since a more significant proportion of the communication volume has been transferred from PCI-e to NVLink.

\subsection{Scaling to multi-node}
\begin{figure}[ht!]
\centering
\includegraphics[width=0.5\textwidth]{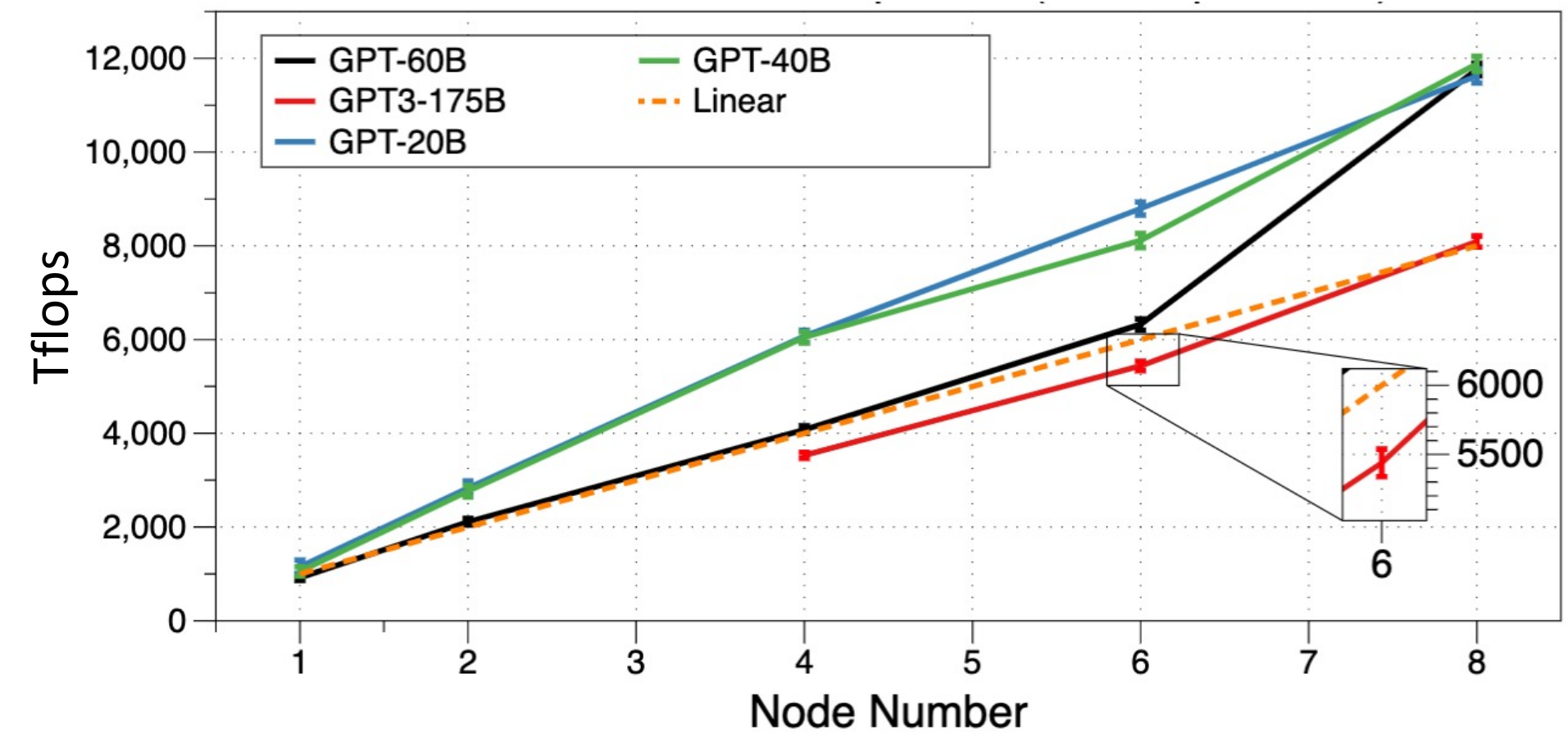}
\caption{
Scalability of PatrickStar on multi-node SuperPod.
}
\label{fig:m_node_scale}
\end{figure}

PartrickStar is also evaluated on a multile node on SuperPod.
Nodes are connected with IB-HAC, whose bi-direction bandwidth is 800 Gb/s.
Figure~\ref{fig:m_node_scale} shows superlinear multi-node scalability for four test cases using batch size per GPU as 8.
We succeed in running a 175B GPT3-scale model on 4 nodes (32 GPUs) and
achieved 7611.5 Tflops on 64 GPUs (38\% of the system's peak performance).

\section{Conclusion}
We proposed an innovative heterogeneous training system called PatrickStar.
It organizes model data into chunks and orchestrates them dynamically in heterogeneous memory space.
The system is symbiotic with the ZeRO-DP.
PatrickStar successfully lowers the hardware requirements for PTM training and scales to multiple GPUs more efficiently.
{On two different GPU clusters, PatrickStar achieves larger trainable model scale (2.27x and 2.5x) and superior efficiency compared to SOTA work.}

\bibliography{my.bib}
\bibliographystyle{ACM-Reference-Format}

\end{document}